\documentclass[prl,nofootinbib,twocolumn,showkeys,groupaddress,preprintnumbers,floatfix,
superscriptaddress]{revtex4-1}

\usepackage{etex}                        \usepackage{ifpdf}                       \usepackage{xspace}                      \usepackage[table]{xcolor}               \usepackage{setspace}                    

\PassOptionsToPackage{final}{graphicx}   \usepackage{graphicx}                    

\makeatletter                            \@ifclassloaded{beamer}{}{\PassOptionsToPackage{pagebackref}{hyperref}}
\makeatother
\usepackage{hyperref}
\usepackage{url}                         \usepackage[super]{nth}                  \usepackage{contour}                     \usepackage{verbatim}                    \usepackage{ragged2e}                    \usepackage{attrib}                      \usepackage{adjustbox}                   \usepackage[absolute,overlay]{textpos}   \usepackage{calc}                        

\usepackage{amsmath}                     \usepackage{amscd}                       \usepackage{amsfonts}                    \usepackage{amssymb}                     \usepackage{amsthm}                      \usepackage{scalerel}                    \usepackage{bm}                          \usepackage{bbm}                         \usepackage{braket}				               \usepackage{mathtools}                   \usepackage{etoolbox}                    \usepackage{textcomp}                    \usepackage{stmaryrd}                    \usepackage{nicefrac}                    \usepackage[binary-units]{siunitx}       

\usepackage[T1]{fontenc}                 \usepackage{lmodern}                     \usepackage[scaled=0.72]{beramono}       \usepackage{microtype}                   

\usepackage{algorithm}                   \usepackage{algorithmicx}                

\usepackage{booktabs}                    \usepackage{dcolumn}                     

\makeatletter
  \@ifclassloaded{revtex4-1}
  {
    \usepackage[caption=false]{subfig}   }
  {
    \usepackage{caption}
    \usepackage{subcaption}
  }
  \@ifclassloaded{beamer}
  {
    \usepackage[backend=biber,
                autocite=superscript,
                doi=false,
                url=false,
                sorting=none]{biblatex}
    \usepackage[useregional,showdow]{datetime2}
  }
  {
\usepackage{enumitem}                }
\makeatother

\definecolor{ryan}{RGB}{64, 0, 64}
\definecolor{nix}{RGB}{255, 0, 0}

\definecolor{ucdblue1}{cmyk}{.87,.46,0,.49} \definecolor{ucdblue2}{cmyk}{1., .56, 0., .34}
\colorlet{ucdblue}{ucdblue2}

\makeatletter
\def\@lox@prtc{\section*{\@fxlistfixmename}\begingroup\def\@dotsep{4.5}}
\def\@lox@psttc{\endgroup}
\makeatother

\usepackage{epigraph}

\usepackage{tikz}
\usetikzlibrary{arrows}
\usetikzlibrary{automata}
\usetikzlibrary{calc}
\usetikzlibrary{decorations.markings}
\usetikzlibrary{decorations.pathreplacing}
\usetikzlibrary{external}
\usetikzlibrary{intersections}
\usetikzlibrary{patterns}
\usetikzlibrary{plotmarks}
\usetikzlibrary{positioning}
\usetikzlibrary{through}

\pgfkeys{/pgf/images/include external/.code=\includegraphics{#1}}

\tikzsetexternalprefix{tikz_cache/}

\tikzstyle{vaucanson}=[
node distance=3cm,
  bend angle=15,
  auto,
every loop/.style={},
  every edge/.style={->,draw=black,line width=1.2,>=latex,shorten <=1pt, shorten >=1pt},
  every state/.style={draw=black,line width=2,font=\large},
  loop right/.style={right,out=22,in=-22,loop},
  loop above/.style={above,out=112,in=68,loop},
  loop left/.style={left,out=202,in=158,loop},
  loop below/.style={below,out=292,in=248,loop},
  loop above right/.style={right,out=67,in=23,loop},
  loop above left/.style={left,out=157,in=113,loop},
  loop below left/.style={left,out=247,in=203,loop},
  loop below right/.style={right,out=337,in=293,loop},
  binode/.style={minimum size=1cm,inner sep=0pt},
]

\tikzset{
  mystate/.style={circle,draw,fill=black}
}

\makeatletter
\newcommand{\binode}[1][state]{\@ifnextchar[{\binode@i[{#1}]}{\binode@i[{#1}][{yellow},{gray!70}]}}
\def\binode@i[#1][#2,#3]{\@ifnextchar({\binode@ii[{#1}][{#2},{#3}]}{\binode@ii[{#1}][{#2},{#3}]({})}}
\def\binode@ii[#1][#2,#3](#4){\@ifnextchar[{\binode@iii[{#1}][{#2},{#3}]({#4})}{\binode@iii[{#1}][{#2},{#3}]({#4})[{}]}}
\def\binode@iii[#1][#2,#3](#4)[#5]#6{

  \node[#1,binode] (#4) [#5] {#6};
  \node[yshift=-10pt] (#4SS) at (#4.270) {};
  \node[xshift=-10pt] (#4WW) at (#4.180) {};
  \node[xshift=-10pt] (#4NN) at (#4.180) {};
  \node[xshift=-10pt,yshift=10pt] (#4NW) at (#4.135) {};
\begin{scope}
    \path[clip] (#4.255) -- +(-.8cm,0cm) -- +(-.8cm,1cm) -- (#4.75) -- cycle;
    \node[#1,fill=#2,binode] (#4f) [#5] {#6};
  \end{scope}
\begin{scope}
    \path[clip] (#4.75) -- +(.8cm,0cm) -- +(.8cm,-1cm) -- (#4.255) -- cycle;
\node[#1,fill=#3,binode] (#4r) [#5] {#6};
  \end{scope}
  \node {} }
\makeatother

\definecolor{FCSA}{RGB}{141,211,199}
\definecolor{FCSB}{RGB}{255,255,179}
\definecolor{RCSC}{RGB}{185,138,196}
\definecolor{RCSD}{RGB}{231,143,111}
\definecolor{RCSE}{RGB}{128,177,211}

\tikzset{
  anticlockwise arc centered at/.style={
    to path={
      let \p1=(\tikztostart), \p2=(\tikztotarget), \p3=(#1) in
      \pgfextra{
        \pgfmathsetmacro{\anglestart}{atan2(\y1-\y3,\x1-\x3)}
        \pgfmathsetmacro{\angletarget}{atan2(\y2-\y3,\x2-\x3)}
        \pgfmathsetmacro{\angletarget}{\angletarget < \anglestart ? \angletarget+360 : \angletarget}
        \pgfmathsetmacro{\radius}{veclen(\x1-\x3,\y1-\y3)}
      }
      arc(\anglestart:\angletarget:\radius pt) -- (\tikztotarget)
    },
  },
  clockwise arc centered at/.style={
    to path={
      let \p1=(\tikztostart), \p2=(\tikztotarget), \p3=(#1) in
      \pgfextra{
        \pgfmathsetmacro{\anglestart}{atan2(\y1-\y3,\x1-\x3)}
        \pgfmathsetmacro{\angletarget}{atan2(\y2-\y3,\x2-\x3)}
        \pgfmathsetmacro{\angletarget}{\angletarget > \anglestart ? \angletarget - 360 : \angletarget}
        \pgfmathsetmacro{\radius}{veclen(\x1-\x3,\y1-\y3)}
      }
      arc(\anglestart:\angletarget:\radius pt)  -- (\tikztotarget)
    },
  },
}

\usepackage{pgfplots}
\usepgfplotslibrary{fillbetween}
\usepgfplotslibrary{groupplots}
\usepgfplotslibrary{units}
\pgfplotsset{compat=newest}

\usepackage{tcolorbox}
\usepackage{circuitikz}

\colorlet {past_color}    {red}
\colorlet {pres_color}    {blue}
\colorlet {futu_color}    {black!30!green}

\colorlet {temp_color_1}  {red!50!blue}
\colorlet {temp_color_2}  {red!50!green}
\colorlet {temp_color_3}  {blue!50!green}

\colorlet {hmu_color}     {blue!67!green}
\colorlet {rhomu_color}   {temp_color_1!80!blue}
\colorlet {rmu_color}     {blue}
\colorlet {bmu_1_color}   {temp_color_1}
\colorlet {bmu_2_color}   {temp_color_3}
\colorlet {qmu_color}     {temp_color_1!67!green}
\colorlet {wmu_color}     {temp_color_2!57!blue}
\colorlet {sigmamu_color} {temp_color_2}

\usepackage{listings}
\lstdefinestyle{mypython}{
language=Python,                        basicstyle=\small\ttfamily,             keywordstyle=\color{green!50!black},    commentstyle=\color{gray},              numbers=left,                           numberstyle=\tiny,                      stepnumber=1,                           numbersep=5pt,                          backgroundcolor=\color{gray!10},        frame=none,                             tabsize=2,                              captionpos=b,                           breaklines=true,                        breakatwhitespace=false,                showspaces=false,                       showtabs=false,                         morekeywords={as},                      }

\theoremstyle{plain}    
\theoremstyle{plain}    
\theoremstyle{plain}    
\theoremstyle{plain}    
\theoremstyle{plain}    
\theoremstyle{plain}    
\theoremstyle{plain}    
\theoremstyle{plain}    
\theoremstyle{plain}    
\theoremstyle{plain}    
\theoremstyle{plain}    
\theoremstyle{plain}

\usepackage[capitalise]{cleveref}        \crefname{Lem}{Lemma}{Lemmas}

\newcommand{\CausalState}       { \mathcal{S} }

\newcommand{\forward}{+}
\newcommand{\reverse}{-}
\newcommand{\forwardreverse}{\pm} 

\newcommand{\FutureCausalState} { {\CausalState}^{\forward} }

\newcommand{\PastCausalState}   { {\CausalState}^{\reverse} }

\newcommand{\lastindex}[2]{
  \edef\tempa{0}
  \edef\tempb{#2}
  \ifx\tempa\tempb
\edef\tempc{#1}
  \else
\edef\tempa{0}
    \edef\tempb{#1}
    \ifx\tempa\tempb
      \edef\tempc{#2}
    \else
      \edef\tempc{#1+#2}
    \fi
  \fi
  \tempc
}

\newcommand{\CSjoint}[1][,]{
   \edef\tempa{:}
   \edef\tempb{#1}
   \ifx\tempa\tempb
\ensuremath{\FutureCausalState\!#1\PastCausalState}
   \else
\ensuremath{\FutureCausalState#1\PastCausalState}
   \fi
}
\newcommand{\CSjointKL}[3][,]{
   \edef\tempa{:}
   \edef\tempb{#1}
   \ifx\tempa\tempb
\ensuremath{\FutureCausalState_{#2}\!#1\PastCausalState_{#3}}
   \else
\ensuremath{\FutureCausalState_{#2}#1\PastCausalState_{#3}}
   \fi
}

\newif\ifpm
\edef\tempa{\forwardreverse}
\edef\tempb{\pm}
\ifx\tempa\tempb
   \pmfalse
\else
   \pmtrue
\fi

\ifcsname mathclap\endcsname
  \relax
\else
  \def\clap#1{\hbox to 0pt{\hss#1\hss}}

\fi

\newcommand{\op} [3] [] {
  \ensuremath{
    \operatorname{#2_{#1}}
    \if\relax\detokenize{#3}\relax
    \else
      \left[ #3 \right]
    \fi
  }
  \xspace
}

\makeatletter
  \@ifclassloaded{beamer}
  {
    \hypersetup{
      linkcolor={blue!50!black},
      citecolor={blue!50!black},
      urlcolor={blue!80!black},
    }
  }
  {
    \hypersetup{
      colorlinks,
      linkcolor={blue!50!black},
      citecolor={blue!50!black},
      urlcolor={blue!80!black},
    }
  }
\makeatother

\usepackage{empheq}
\usepackage{overpic}

\newcommand{\ones}{\vec{1}}

\makeatletter
\@booleanfalse\titlepage@sw
\makeatother

\begin{document}

\def\ourTitle{Geometry and Dynamics of LayerNorm
}

\def\ourAbstract{A technical note aiming to offer deeper intuition for the LayerNorm function common in deep neural networks.
LayerNorm is defined relative to a distinguished `neural' basis, but it does more than just normalize the
corresponding vector
elements.
Rather, it implements a composition---of linear projection, nonlinear scaling, and then affine transformation---on
input activation vectors.
We develop both a new mathematical expression and geometric intuition, to make the net effect more transparent.
We emphasize that, when LayerNorm acts on an $N$-dimensional vector space, 
all outcomes of LayerNorm lie within the intersection of an $(N \! - \!1)$-dimensional hyperplane and 
the interior of
an $N$-dimensional hyperellipsoid.
This intersection is the interior of an  
$(N \! - \!1)$-dimensional hyperellipsoid,
and
typical inputs are mapped near 
its surface.
We find the direction and length of the principal axes 
of this $(N \! - \!1)$-dimensional hyperellipsoid
via the eigen-decomposition of a simply constructed matrix.
}

\def\ourKeywords{}

\hypersetup{
  pdfauthor={Paul M. Riechers},
  pdftitle={\ourTitle},
  pdfsubject={\ourAbstract},
  pdfkeywords={\ourKeywords},
  pdfproducer={},
  pdfcreator={}
}

\title{\ourTitle}

\author{Paul M. Riechers}
\email{pmriechers@gmail.com}

\affiliation{Simplex AI Safety,
	Berkeley, CA}

\affiliation{Beyond Institute for Theoretical Science (BITS),
San Francisco, CA}

\date{\today}
\bibliographystyle{unsrt}

\begin{abstract}
\ourAbstract
\end{abstract}

\date{\today}
\maketitle

\setstretch{1.1}

\section{Introduction}

LayerNorm is a relatively simple function and so is often taken for granted.  But, as one of the few building blocks of modern transformers (and other deep neural networks), it is worth understanding deeply.
At the risk of stating things that are obvious, this note aims to very explicitly think through the LayerNorm function, how it maps activations, and the resultant geometry of activations.

LayerNorm acts on each $N$-dimensional activation vector $\vec{a} \in \mathbb{R}^N$ via a nonlinear transformation induced by learned parameters
$\vec{g}$, $\vec{b} \in \mathbb{R}^N$
and fixed small parameter $\epsilon \in \mathbb{R}_{\geq 0}$~\cite{Ba16_Layer, PyTorchLayerNorm}.
Let $\ones$ denote the $N$-dimensional vector of all ones in the neural basis.
Then LayerNorm can be written as 
\begin{align}
	\text{LayerNorm}(\vec{a}, \vec{g}, \vec{b}, \epsilon) = 
	\vec{g} \odot \frac{(\vec{a} - \mu \ones)}{\sqrt{\sigma^2+\epsilon}} + \vec{b} ~.
\end{align}	
where $\odot$ denotes the element-wise product (i.e., Hadamard product), the mean neural activation is
\begin{align}
\mu = \vec{a} \cdot \ones / N
\end{align}	
and the variance of activations across neurons is 
\begin{align}
\sigma^2 = (\vec{a} - \mu \ones) \cdot (\vec{a} - \mu \ones)  / N ~.
\end{align}	
Typically, the gain $\vec{g}$ and bias $\vec{b}$ are learned during training.
At the beginning of training, $\vec{g} = \ones$ and $\vec{b} = \vec{0}$, and $\epsilon = 10^{-5}$ by default.
This can alternatively be written in standard matrix-multiplication notation as
\begin{align}
	\text{LayerNorm}(\vec{a}, \vec{g}, \vec{b}, \epsilon) = 
	\text{diag}(\vec{g}) \, \frac{(\vec{a} - \mu \ones)}{\sqrt{\sigma^2+\epsilon}} + \vec{b} ~,
\end{align}	
where diag$(\vec{g})$ is the matrix of all zeros except for the elements of $\vec{g}$ along its diagonal.
At the beginning of training, diag$(\vec{g})$ is the identity matrix, $I$.

\section{Decomposing and Visualizing LayerNorm}

\begin{figure*}[b]
\includegraphics[width=\textwidth]{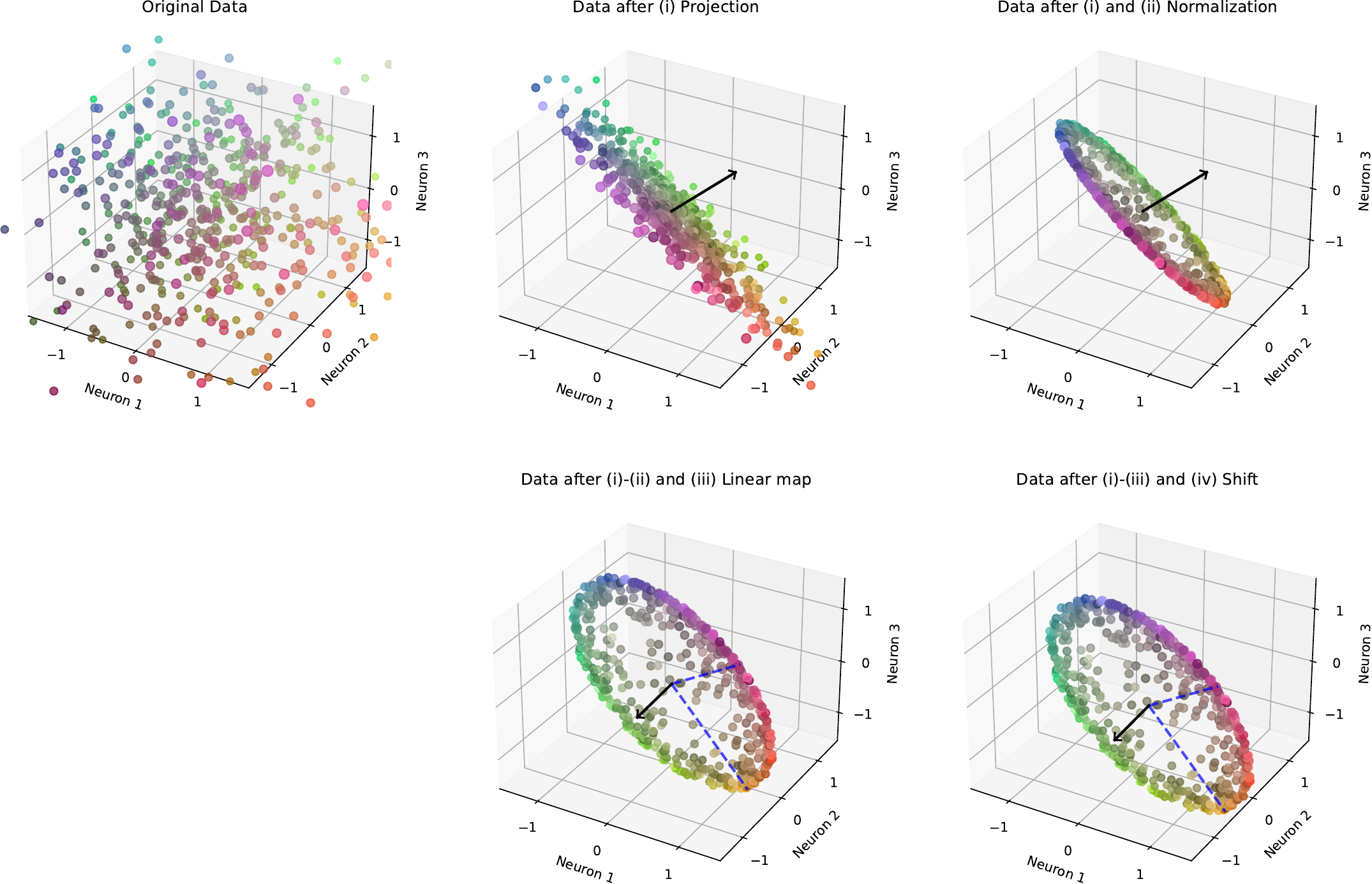}
\caption{LayerNorm as a composition of (i) Projection, (ii) Normalization, (iii) Linear transformation, and (iv) Global shift.  
Each point in the sequence of panels represents an activation vector input to LayerNorm.  The (R,G,B) color values 
of each point directly encode the original activation of input neurons 1, 2, and 3, respectively in this simple example with $N=3$.
The second and third panels explicitly show the $\hat{1}$ vector orthogonal to the initial projection, while the last two panels show 
the unit vector $\hat{\alpha} = \vec{\alpha} / \alpha$ (with neural-basis components of $\vec{\alpha}$ given by $\alpha_n = 1/g_n$) which is orthogonal to the 
new plane after the linear transformation by diag$(\vec{g})$.
(The arrow depicting $\hat{\alpha}$ is shifted by $\vec{b}$ in the final case.)
The last two panels also show the semi-axes of the principal axes as dashed blue lines, which correspond to
eigenstates of $\Pi_2 G^{-2} \Pi_2$, with length $\sqrt{N/\lambda}$.
We have depicted an unusually large $\epsilon = 1/10$ to demonstrate that the interior of the hyperellipsoid is not strictly empty after LayerNorm.  But see Fig.~2.}
\label{fig:LN_steps}
\end{figure*}

LayerNorm's net
action on the
$N$-dimensional activation vector $\vec{a}$ 
can be thought of as a composition of several other simple functions (i)-(iv), as depicted in Fig.~\ref{fig:LN_steps}.  
First, LayerNorm performs a 
linear projection: \begin{itemize}
	\item[(i)]
	Find the mean value of components in the neural basis, and subtract this mean from each component.
	Notably $\vec{a}' = \vec{a} - \mu \ones =  \vec{a} - (\vec{a} \cdot \hat{1} ) \hat{1} = (I - \hat{1} \hat{1}^\top) \vec{a}$ can be seen as the projection of $\vec{a}$ onto the $(N \! - \!1)$-dimensional  
	hyperplane perpendicular to $\ones$.
\end{itemize}
Here, $\hat{1} = \ones / \sqrt{N}$ is the unit vector in the direction of $\ones$.
Note that $\vec{a}' \cdot \hat{1} = \vec{0}$
since 
$\vec{a}' \cdot \hat{1} 
= \bigl[  \vec{a} - (\vec{a} \cdot \hat{1} ) \hat{1} \bigr] \cdot \hat{1} 
=  \vec{a} \cdot \hat{1} - (\vec{a} \cdot \hat{1} ) \hat{1} \cdot \hat{1} 
= \vec{a} \cdot \hat{1} -\vec{a} \cdot \hat{1} = \vec{0}
$.
As a consequence, this initial sub-step of LayerNorm
projects the activation vector onto a hyperplane
	that enforces $\sum_{n} a_n' = 0$,
	where $a_n'$ is the $n^\text{th}$ component of the resultant vector $\vec{a}'$ in the neural basis.

Let's denote the rank-$(N \! - \! 1)$ projector $\Pi = I - \hat{1} \hat{1}^\top$.
Then we can write the result of sub-step (i) simply as 
$\vec{a}' = \Pi \vec{a}$.
Note that, as a projector, $\Pi^2 = \Pi$.
Hence, 
the variance can either be expressed as
$\sigma^2 = \vec{a}^\top \Pi^\top \Pi \vec{a} /N= \vec{a}^\top \Pi \vec{a} /N$,
or in relation to the square magnitude of $\vec{a}'$ since
$\sigma^2 = |\Pi \vec{a}|^2/N$.
	
Subsequently, LayerNorm performs a nonlinear scaling:	
\begin{itemize}	
	\item[(ii)] 
	Scale the magnitude of the resultant vector 
	by $(\sigma^2 + \epsilon)^{-1/2}$,
	after which it is
no larger than $\sqrt{N}$.
	This scales the vector to bring it within the $N$-ball.
	If $\epsilon$ is small, then resultant points will be concentrated towards a magnitude of $\sqrt{N}$.
\end{itemize}

After the first two sub-steps (i) and (ii), the 
activations have been projected and nonlinearly scaled
according to:
\begin{align}
\vec{a} \mapsto 
\begin{cases}
	\vec{0} & \text{if } \vec{a} \propto \ones \\
\frac{\sqrt{N} \, \widehat{\Pi \vec{a}}  }{\sqrt{1 + N \epsilon / \vec{a}^\top \Pi \vec{a} }} & \text{otherwise}
\end{cases}
~.
\end{align}	
$ \widehat{\Pi \vec{a}} $ is a unit vector in the direction of $\Pi \vec{a}$, while the denominator $\sqrt{1 + N \epsilon / \vec{a}^\top \Pi \vec{a} }$ is never smaller than 1.
Accordingly, after these first two sub-steps, 
every input activation vector is
mapped into the intersection of (a)
the 
$(N \! - \!1)$-dimensional  
hyperplane perpendicular to $\ones$
and (b) the $N$-ball of radius $N^{1/2}$.

After these first two sub-steps, 
the $N$-sphere and the points it now contains are subsequently (iii) stretched by $\vec{g}$ along the neural-basis directions into a hyperellipsoid, and then (iv) shifted by $\vec{b}$. 
Together, sub-steps (iii) and (iv) constitute
an affine transformation. The linear transformation (iii)---matrix multiplication by diag$(\vec{g})$---is diagonal in the neural basis, and maps the open $N$-ball containing all points to the interior of an $N$-dimensional hyperellipsoid with principal axes given by the neural basis.
Likewise, it stretches and so tilts the intersecting hyperplane that contains all relevant points. 
The bias (iv) simply shifts the entire hyperellipsoid by $\vec{b}$.

We can now combine all these sub-steps to re-express 
LayerNorm as
\begin{align}
	\text{LayerNorm}(\vec{a}, \vec{g}, \vec{b}, \epsilon) = 
	\sqrt{N} \text{diag}(\vec{g}) \, \frac{ \Pi \vec{a} }{\sqrt{| \Pi \vec{a}|^2+ N \epsilon }} + \vec{b} ~,
	\label{eq:LayerNormNewEmph}
\end{align}	
where we have defined the projector
$\Pi = I - \hat{1} \hat{1}^\top$.
This expression has the advantage of making all $\vec{a}$-dependence explicit, while highlighting the projection, scaling, and subsequent affine transformation.

The sequence of sub-steps building up the dynamics and geometry of LayerNorm is depicted in Fig.~\ref{fig:LN_steps}.
Although LayerNorm is usually used in high-dimensional vector spaces, it is instructive to visualize and understand how it would 
operate in three dimensions, as shown, from which higher-dimensional behavior can be 
both intuitively and rigorously extrapolated.

\begin{figure*}[t]
	\includegraphics[width=\textwidth]{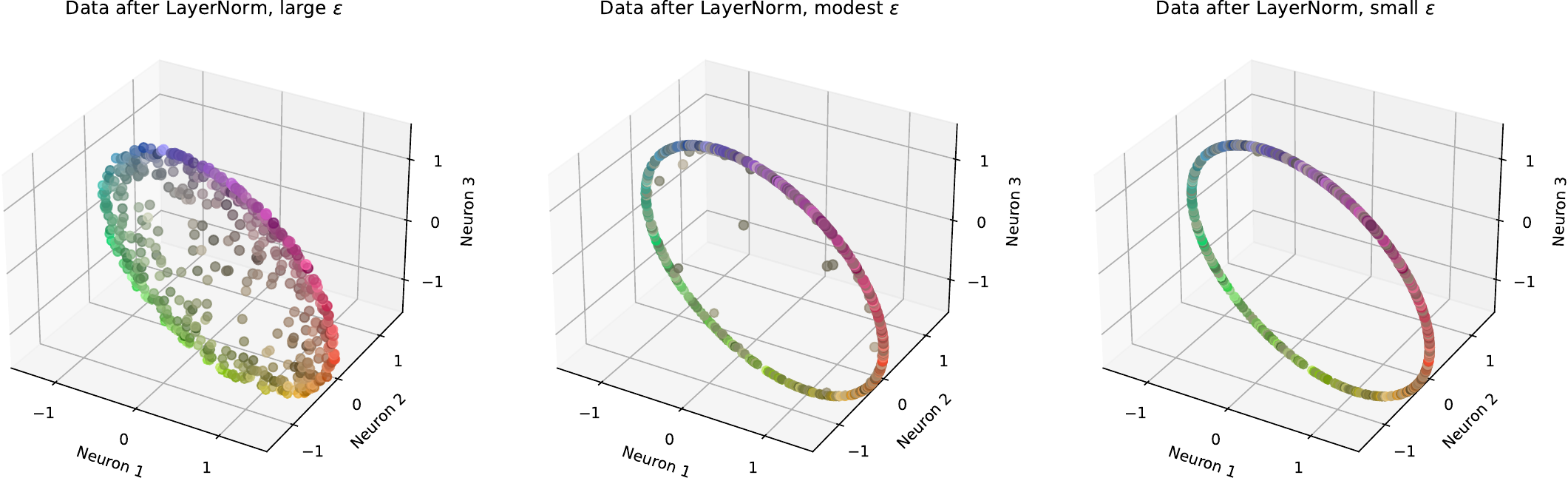}
	\caption{From left to right, we compare the net effect of LayerNorm for $\epsilon = 10^{-1}$, $10^{-3}$, and $10^{-5}$. The final case corresponds to the default value for this small parameter in the standard PyTorch LayerNorm function.  In practice, we should expect that most points get mapped very near the surface of the hyperellipsoid.
}
\label{fig:eps_comparison}
\end{figure*}

In particular, 
when LayerNorm acts on an $N$-dimensional vector space, 
all outcomes of LayerNorm lie within the intersection of an $(N \! - \! 1)$-dimensional hyperplane and an $N$-dimensional hyperellipsoid.  
This intersection is itself a hyperellipsoid, now of dimension $N-1$.
After a sprinkling of points is transformed by LayerNorm, 
most of the points subsequently contained in this 
$(N \! - \! 1)$-dimensional
hyperellipsoid are concentrated towards its surface, 
as shown in Fig.~\ref{fig:eps_comparison}.

\subsection{Orthogonal subspace}

If we subtract the bias, then the image of all activations after LayerNorm is orthogonal to at least one dimension of the vector space.  This orthogonal space $\mathcal{N}$ can be found as the left nullspace of $\text{diag}(\vec{g}) - \vec{g} \ones^\top / N$.
Let $\{ \ket{n} \}_{n=1}^N$ denote the orthonormal neural basis.
If the elements of $\vec{g}$ are all non-zero in the neural basis---i.e., if $g_n = \vec{g}^\top \ket{n} \neq 0$ for any $n$---then we find that
the orthogonal subspace is the one-dimensional subspace of
all vectors proportional to the inverse elements of $\vec{g}$.  I.e., 
$\mathcal{N} = \{ c \vec{\alpha} : c \in \mathbb{R}, \text{ and } \alpha_n = \vec{\alpha}^\top \ket{n} = 1/g_n  \}$.
However, if $g_n = 0$ for at least one $n$, 
then the number of orthogonal directions 
is equal to the number of these elements equal to zero,
with the orthogonal subspace now given instead by
$\mathcal{N} = \text{span} \bigl( \{ \ket{n} : g_n = 0 \} \bigr)$.

\subsection{Principal axes}

The neural basis forms the principal
axes of the $N$-dimensional hyperellipsoid 
after the $N$-sphere is stretched by $\vec{g}$.
But what are the principal axes of the 
cross-sectional $(N \! - \! 1)$-hyperellipsoid 
induced by $\vec{g}$?
This is a rather more difficult question.

We find that the principal axes are the eigenstates of 
$\Pi_2 G^{-2} \Pi_2$, 
with semi-axes of length $\sqrt{N/\lambda}$ 
where $\lambda$ is the eigenvalue associated with the eigenstate, $G = \text{diag}(\vec{g})$, and $\Pi_2 = I - \vec{\alpha} \vec{\alpha}^\top / \vec{\alpha}^\top \vec{\alpha}$.
As above, $\vec{\alpha}$ contains the reciprocal elements of the scaling vector in the neural basis:
$\alpha_n = \vec{\alpha}^\top \ket{n} = 1/g_n$.

If $\vec{g}$ has neural-basis components equal to zero, i.e., if $g_n = 0$ for some $n$, then $G^{-2}$ should be interpreted as the square of $G$'s Drazin inverse~\cite{Riec18_Beyond}, while
we can then set either $\Pi_2 = I$ or $\Pi_2 = I - \ket{n} \! \bra{n}$, with $\braket{n | m} = \delta_{n,m}$.
If $g_n \neq 0$ for any $n$, then $\Pi_2 G^{-2} \Pi_2$
has a single zero eigenvalue, with corresponding eigenstate $\vec{\alpha}$ in the subspace orthogonal
to LayerNorm's image.

Lengths of the semi-axes can alternatively be found as $\sqrt{N \zeta}$ where each $\zeta$ is an eigenvalue of $\Pi G^2 \Pi$.  However, unlike the use of $\Pi_2 G^{-2} \Pi_2$ above, this does not deliver the directions of the principal axes.

\section{LayerNorm in transformers}

In the standard modern transformer architecture, 
LayerNorm is invoked in two or three different ways, depending on how you count.
It is used outside of the residual stream, both before multi-head 
attention and before the position-wise feed-forward network~\cite{Xiong20_Layer}.
And it is used in the final residual stream, just before unembedding.

\section{Conclusion}

We have investigated LayerNorm
as a composition of simpler functions---projection, scaling, and then affine transformation---and have derived 
an alternative expression (Eq.~\eqref{eq:LayerNormNewEmph}) for LayerNorm that makes these features more evident.
Some work had already been done in this direction---e.g., the projection sub-step implied by LayerNorm was already noted in Refs.~\cite{Brody23_Expressivity, Molina23_Traveling}.
Our short note
provides complementary 
perspective.  We included the sometimes non-negligible effect of the small $\epsilon$ parameter used in the standard PyTorch implementation of LayerNorm.  We have also identified the orthogonal subspace to the bias-corrected image of all activations after LayerNorm, which may be useful in understanding how LayerNorm interacts with downstream components of a neural net.
Finally, we have identified the principal axes (and their lengths) of the $(N \! - \! 1)$-dimensional 
hyperellipsoid image of LayerNorm via the eigendecomposition of a simply constructed matrix ($\Pi_2 G^{-2} \Pi_2$) that we introduced here.

Better understanding the components of neural networks
should help us anticipate their implications, both locally and in composition with other aspects of the network.  LayerNorm is a ubiquitous function used in modern neural nets, 
with more complexity than it initially leads on.  
We hope this exposition makes it more intuitive, and brings 
at least 
another pinhole of light into the black box.

\end{document}